%% file: main.tex
\documentclass[10pt,twocolumn,letterpaper]{article}

\usepackage{cvpr}              % To produce the CAMERA-READY version
\usepackage{csquotes}
\usepackage{algorithm}
\usepackage{algpseudocode}
\usepackage{multirow}

\usepackage{textcomp} 
\usepackage{marvosym}
\usepackage{colortbl}
\usepackage{xcolor}
\usepackage{diagbox}
\usepackage{wasysym}
\usepackage{marvosym}
\usepackage{pifont}
\usepackage{setspace}

\usepackage{enumitem}

\usepackage{makecell}
\input{preamble}

\definecolor{cvprblue}{rgb}{0.21,0.49,0.74}
\usepackage[pagebackref,breaklinks,colorlinks,allcolors=cvprblue]{hyperref}
\usepackage[most,theorems,skins,listings]{tcolorbox}

\def\paperID{5447} % *** Enter the Paper ID here
\def\confName{CVPR}
\def\confYear{2026}

\title{Group Editing\adjustbox{valign=c}{\includegraphics[height=2.4em]{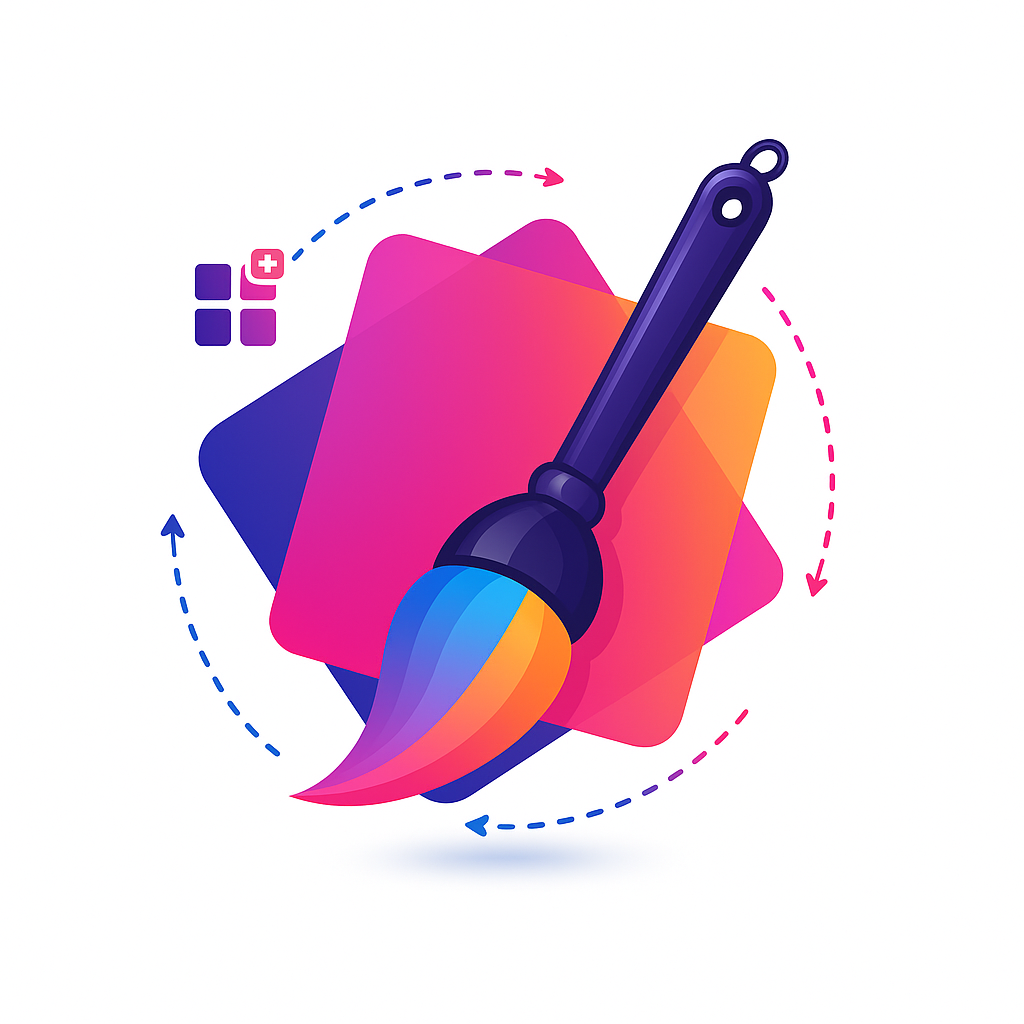}}:   Edit Multiple Images in One Go }
\author{%
  Yue Ma\textsuperscript{1\dag},
  Xinyu Wang\textsuperscript{2\dag},
  Qianli Ma\textsuperscript{3\S},
  Qinghe Wang\textsuperscript{\textbf{\Letter}},
  Mingzhe Zheng\textsuperscript{1},
  Xiangpeng Yang\textsuperscript{4},\\
  Hao Li\textsuperscript{2},
  Chongbo Zhao\textsuperscript{2},
  Jixuan Ying\textsuperscript{2},
  Harry Yang\textsuperscript{1},
  Hongyu Liu\textsuperscript{1}\textsuperscript{\textbf{\Letter}},
  Qifeng Chen\textsuperscript{1}
  \\[1mm]
  \textsuperscript{1} HKUST \quad 
  \textsuperscript{2} THU \quad
  \textsuperscript{3} SJTU  \quad
  \textsuperscript{4} University of Technology Sydney
   \\
  \\
  \textbf{Project: \href{https://group-editing.github.io/}{\texttt{\textcolor{cyan}{https://group-editing.github.io/}}}}
}

\begin{document}
% \maketitle

\twocolumn[{
\begin{center}
\maketitle
\vspace{-2em}
    \captionsetup{type=figure}
    \includegraphics[width=0.85\textwidth]{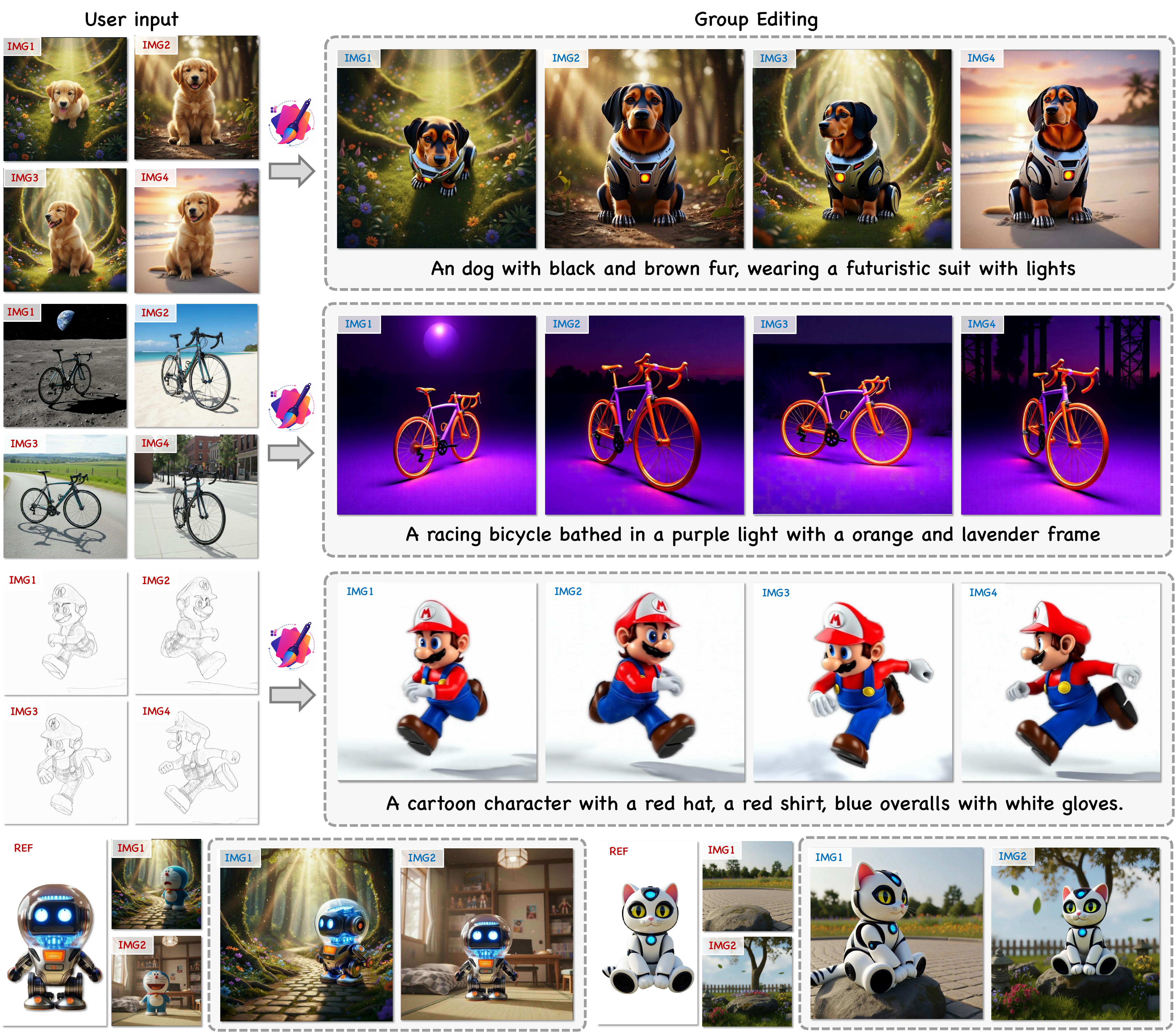}
    \vspace{-0.5em}
    \caption{\textbf{Gallery of proposed method.} We propose the~\modelname, which aims to apply consistent and unified modifications across a set of related images. Our~\modelname supports local and global editing, colorization, customization, and insertion among image groups.}
\end{center}
}]

\begingroup
\renewcommand{\thefootnote}{}
\footnotetext{\dag~Equal contribution.}
\footnotetext{\Letter~Corresponding author.}
\footnotetext{\S\ Project leader}
\endgroup

\clearpage

\input{sec/0_abstract}    
\input{sec/1_introduction}
\input{sec/2_related}
\input{sec/3_method_V4}

\input{sec/4_experiment}

\input{sec/5_conclusion}
\input{sec/X_suppl}
{
    \small
    \bibliographystyle{ieeenat_fullname}
    \bibliography{main}
}


\end{document}

%% file: preamble.tex
\newcommand{\modelname}{\textit{GroupEditing}\xspace}
\definecolor{myorange}{RGB}{255,100,3}
\definecolor{mygray}{gray}{.85}
\definecolor{mygray1}{gray}{.7}
\definecolor{mygray2}{gray}{.93}
\definecolor{mygray3}{gray}{.90}
\usepackage{tikz}
\usepackage{array}
\usepackage{graphicx}      % for resizing the table (resizebox)
\usepackage{multirow}      % for multi-row cells in the table
\usepackage{colortbl}      % for coloring the cells
\usepackage{booktabs}      % for better table formatting with \toprule, \midrule, and \bottomrule
\usepackage{xcolor}        % for text color (like \textcolor)
\usepackage{caption}       % for better control of captions (optional)
\usepackage{gensymb}
\newcolumntype{C}[1]{>{\centering\arraybackslash}p{#1}}

\usepackage{pifont}
\definecolor{iccvblue}{rgb}{0.21,0.49,0.74}
\definecolor{Blue}{RGB}{12, 114, 186} 
\usepackage{color, xcolor}
\usepackage{svg}
\usepackage{pifont}
\usepackage{graphicx}
\usepackage{adjustbox}
\usepackage{algorithm}
\usepackage{algpseudocode}

%% file: sec/0_abstract.tex
\begin{abstract}
In this paper, we tackle the problem of performing consistent and unified modifications across a set of related images. This task is particularly challenging because these images may vary significantly in pose, viewpoint, and spatial layout. Achieving coherent edits requires establishing reliable correspondences across the images, so that modifications can be applied accurately to semantically aligned regions. To address this, we propose GroupEditing, a novel framework that builds both explicit and implicit relationships among images within a group. On the explicit side, we extract geometric correspondences using VGGT, which provides spatial alignment based on visual features. On the implicit side, we reformulate the image group as a pseudo-video and leverage the temporal coherence priors learned by pre-trained video models to capture latent relationships. To effectively fuse these two types of correspondences, we inject the explicit geometric cues from VGGT into the video model through a novel fusion mechanism. To support large-scale training, we construct GroupEditData, a new dataset containing high-quality masks and detailed captions for numerous image groups. Furthermore, to ensure identity preservation during editing, we introduce an alignment-enhanced RoPE module, which improves the model’s ability to maintain consistent appearance across multiple images. Finally, we present GroupEditBench, a dedicated benchmark designed to evaluate the effectiveness of group-level image editing. Extensive experiments demonstrate that GroupEditing significantly outperforms existing methods in terms of visual quality, cross-view consistency, and semantic alignment.
\end{abstract}

%% file: sec/1_introduction.tex
\section{Introduction}
\label{sec:intro}

Group-image editing aims to achieve consistent and unified content creation across a set of related images. Unlike single-image editing~\cite{tumanyan2023plug,mokady2023null,chen2024zero,zhang2023adding,brooks2023instructpix2pix,mou2024t2i}, which only focuses on generating plausible results for an individual image, this task requires maintaining editing coherence among multiple images in both appearance and structure. It is crucial in many emerging applications. For example, in virtual content creation, maintaining coherent edits across images ensures identity integrity for digital avatars and characters~\cite{liu2023facechain,avrahami2024chosen}.
In digital commerce, consistent product depiction across angles improves consumer trust and supports robust visual recommendations. Moreover, this task also improves both the efficiency and quality of downstream tasks such as photo retouching~\cite{chen2018deep}, data augmentation for personalization~\cite{ruiz2023dreambooth}, and 3D reconstruction~\cite{kerbl20233d,wang2025vggt}.

While consistent editing is highly desirable, ensuring coherence across diverse images remains a challenge. Previous editing approaches typically operate on a per-image basis, resulting in variations that undermine the uniformity necessary for practical applications~\cite{mou2024t2i,zhang2023adding, ma2024followyouremoji, ma2025followyourclick, chen2025contextflow, zhu2025multibooth, zhu2024instantswap, wang2024cove, brooks2023instructpix2pix,chen2024zero,mokady2023null,yang2023paint}. 
Optimization-based frameworks~\cite{ju2024brushnet,zhang2023adding,yang2023paint} attempt to edit one image and propagate the modifications across images, but the lack of robust generalization leads to artifacts and inconsistency. Recent optimization-free methods~\cite{bai2025edicho} rely on semantic correspondences from attention features and open-sourced tracking tools, but their application is limited to a small number of images.
A major challenge stems from the lack of high-quality training pairs and sufficient constraints to maintain uniformity, especially in geometrically complex scenes (\textit{e.g.}, identifying the 'left eye' in various images or tracking a logo on a T-shirt during a 30-degree rotation), which hinders generalization.
% They lack generalization due to the lack of high-quality training pair data and the inability to impose sufficient constraints to preserve uniformity across images, particularly in geometrically complex scenes (\textit{e.g.}, identifying \enquote{left eye} in different images, or tracking a specific logo on a T-shirt across a 30-degree rotation).

In this paper, we revisit the problem of group-image editing and explore the question: \textit{What is an appropriate representation for establishing correspondence in this task?} We approach this question from two key observations. 
{\bf i)} \textbf{Implicit correspondence}: Compared to image models, video models inherently possess stronger generative priors and temporal coherence. Trained on large-scale sequential data, video models learn to capture not only semantic continuity but also the spatial transformations of objects. This implicit correspondence allows models to maintain both temporal and spatial consistency naturally, making video models a promising foundation for consistent group-image editing. {\bf ii)} \textbf{Explicit correspondence}: We observe that the semantic correspondence in the video model alone is insufficient. While attention-based matching or pre-trained feature correspondences can align semantic regions (\textit{e.g.}, \enquote{face to face}, \enquote{object to object}), they often fail in geometrically complex scenes where local structures undergo rotation, deformation, or occlusion. To migrate this, we leverage explicit correspondences extracted from the VGGT~\cite{wang2025vggt}, which offers dense matching quality, exemplifying the advantage of strong feature representations.

Motivated by the observations above, we propose GroupEditing, a novel training-based framework for consistent group-image editing. Instead of treating each image independently, we reformulate a set of related images as \textit{pseudo video frames}, enabling the model to inherit the temporal and geometric priors learned by large-scale video models. 
The explicit correspondence representation is obtained from the VGGT~\cite{wang2025vggt} encoder and injected into the video model~\cite{wan2025} with the proposed fine-grained geometry-enhanced RoPE~(Ge-RoPE). The identity-enhanced RoPE is designed for consistent identity preservation. Additionally, to enable large-scale training, we develop a scalable dataset pipeline leveraging various advanced vision models~\cite{kirillov2023segment,liu2024grounding,wang2024qwen2,comanici2025gemini}, constructing the largest group-image editing dataset, \textit{GroupEditData}, and benchmark, \textit{GroupEditBench},  with over 800 groups of images, each featuring precise segmentation masks and detailed text descriptions. We further demonstrate the potential of \modelname by establishing an inpainting-based image editing pipeline that delivers promising results.
To evaluate the effectiveness of our approach, we compare \modelname with prior state-of-the-art (SOTA) baselines. \modelname consistently outperforms all competitors, achieving superior visual quality. In summary, our primary contributions are:

\begin{itemize}
    \item We tackle the problem of performing consistent multiple image editing, and propose \modelname, the first training-based framework that reformulates a set of related image sequences as pseudo video frames.
    \item To enable consistent group-image editing, we introduce geometry-enhanced RoPE that integrates both implicit and explicit correspondence, along with an identity-enhanced RoPE module for robust identity alignment.
    \item To train our model, we develop a dataset pipeline, construct GroupEditData ($>$ 7K groups) and GroupEditBench, each with precise masks and detailed captions.
    \item Experiments show that \modelname achieves SOTA performance across four metrics, including visual quality, editing consistency, and semantic alignment, outperforming baselines on the GroupEditBench.
\end{itemize}

%% file: sec/2_related.tex
\section{Related Work}
\label{sec:related_work}

%-------------------------------------------------------------------------
\noindent\textbf{Generative models for image editing.}
Image editing, a long‐standing problem with broad practical impact, has been substantially advanced by generative modeling~\cite{kingma2013auto,goodfellow2014generative,ho2020denoising,huang2025diffusion,rombach2022high,avrahami2022blended,kim2022diffusionclip, ma2024followpose, ma2025followcreation, ma2026fastvmt, ma2025followyourmotion}.
Recent advances in diffusion models have enabled highly effective editing paradigms~\cite{brooks2023instructpix2pix,tumanyan2023plug,cao2023masactrl,mokady2023null,chen2024zero,feng2025dit4edit,wang2024taming,objectmover,yang2025unified,lan2025efficient}.
Existing techniques can be broadly divided into inference‐time zero‐shot methods that edit images by manipulating the diffusion process itself~\cite{tumanyan2023plug,mokady2023null,cao2023masactrl} and training‐based methods, which achieve editing by fine‐tuning latent diffusion models~\cite{brooks2023instructpix2pix,zhang2023adding,mou2024t2i,patashnik2024consolidating}.
However, these methods remain tailored to single-image editing and, despite efforts to enforce consistency~\cite{winter2024objectdrop,mokady2023null,cao2023masactrl}, they are constrained to small inputs and often break down under complex geometric variation, which is partly due to scarce paired training data. To address this, we formalize \textit{Group-Image Editing} and introduce \textit{\modelname}, a framework that inherits consistency priors for reliable alignment.

\noindent\textbf{Video prior for editing task.}
Video generative models~\cite{ma2025controllable, ma2025followfaster,  ma2022visual, winter2024objectdrop,ma2025decouple,yang2025unified, bai2025edicho, long2025follow, shen2025follow} offer strong temporal consistency priors that can be effectively transferred to image editing.
Existing explorations fall into two main directions: utilizing video data for training data curation~\cite{bagel,chen2025unireal,xiao2025omnigen} and leveraging video models for inference-time guidance~\cite{rotstein2025pathways,wu2025chronoedit}.
While effective for enhancing single-image quality or ensuring short-range consistency, these methods do not solve the fundamental challenge of \textit{Group-Image Editing} across diverse static views.
% Our approach differs by re-framing the image group as a pseudo video sequence, allowing us to explicitly inherit the powerful spatio-temporal coherence and geometric priors of large-scale video models for robustly unified editing.

%% file: sec/3_method_V4.tex
\section{Method}
\label{sec:method}

\begin{figure}[t]
    \centering
    \includegraphics[width=0.9\linewidth]{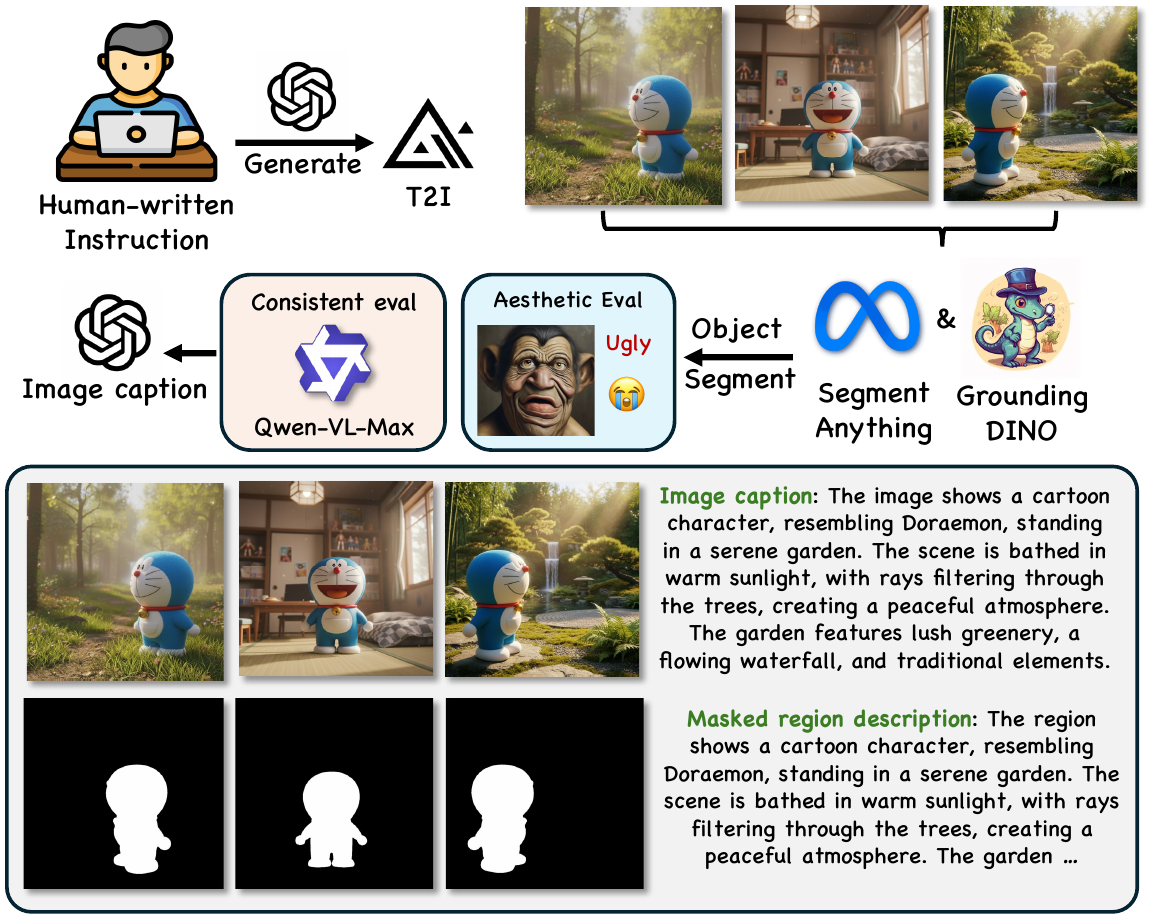}
    \caption{\textbf{Data curation pipeline.} Our pipeline processes human-written instructions through text-to-image generation, followed by quality evaluation (consistency and aesthetic assessment) and annotation generation (object segmentation and text description). The pipeline automatically produces high-quality training pairs with precise masks and detailed captions.}
    \label{fig:data_pipeline}
\end{figure}

\begin{figure*}[ht]
    \centering
    \includegraphics[width=0.95\textwidth]{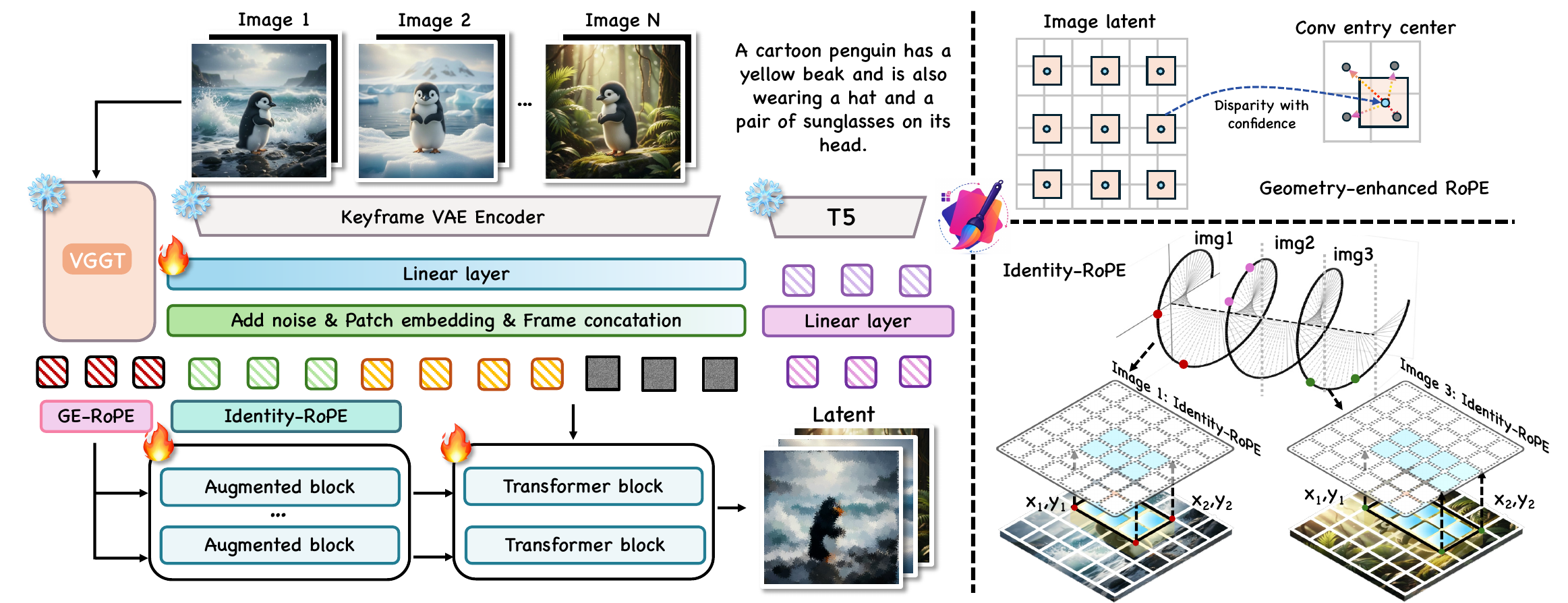}
    \caption{\textbf{Overview of proposed method}. 
    Given a series of images and their corresponding masks, we propose a novel framework for editing while ensuring the consistency of multiple images. To achieve fine-grained spatial alignment, we introduce Geometry-enhanced RoPE (GE-RoPE), which enhances the model’s ability to maintain consistent spatial relationships across different frames, and Identity-RoPE for better consistent identity preservation.
    }
    \label{fig:qualitative-intro}
\end{figure*}

% We present \textbf{GroupEditing}, a training-based framework for consistent group-image editing that reformulates a set of related images as pseudo-temporal frames. The framework integrates implicit spatio-temporal priors from a large video backbone with explicit geometric correspondence, enabled by two specialized RoPE~\cite{su2024roformer} mechanisms: \textbf{Ge-RoPE} for domain alignment and \textbf{Identity-RoPE} for identity preservation. An overview is shown in Fig.~\ref{fig:qualitative-intro}.

We present \textbf{GroupEditing}, a training-based framework for consistent group-image editing that reformulates a set of related images as pseudo-temporal frames. To support large-scale training, we develop a data curation pipeline that automatically generates high-quality training pairs, described in Sec.~\ref{subsec:data}. 
% The framework integrates implicit spatio-temporal priors from a large video backbone with explicit geometric correspondence from VGGT.
% The pipeline processes input images through latent encoding, positional encoding, and explicit token fusion, as detailed in Sec.~\ref{subsubsec:pipeline}. 
For fine-grained spatial alignment, we introduce \textbf{Ge-RoPE} to improve alignment between VGGT features and latent features by incorporating spatial disparity information with confidence levels (See  Sec.~\ref{subsubsec:ge-rope}). For identity preservation, we propose \textbf{Identity-RoPE} to maintain identity consistency across the group through precise pixel-level alignment within each image (Sec.~\ref{subsubsec:identity-rope}). An overview of the framework is illustrated in Fig.~\ref{fig:qualitative-intro}.

\subsection{Data curation}
\label{subsec:data}

To facilitate large-scale training of GroupEditing, we develop a scalable data curation pipeline that automatically generates high-quality training pairs with precise masks and detailed text descriptions. The pipeline, illustrated in Fig.~\ref{fig:data_pipeline}, consists of three main stages: image generation, quality evaluation, and annotation generation.

\subsubsection{Group image generation}
Given a set of human-written text instructions $\mathcal{C}=\{c^{(n)}\}_{n=1}^N$, we employ Gemini 2.5~\cite{comanici2025gemini} as the text-to-image (T2I) generation model to produce corresponding image sets $\mathcal{I}=\{I^{(n)}\}_{n=1}^N$, where each $I^{(n)}=\{I_t^{(n)}\}_{t=1}^{T^{(n)}}$ represents a group of related images generated from the same instruction $c^{(n)}$. The T2I model generates diverse images that share semantic content while exhibiting variations in viewpoint, pose, or scene composition, naturally forming pseudo-temporal sequences suitable for group-image editing training. In total, 18248 instruction–image groups are generated for subsequent processing.

\subsubsection{Quality evaluation}
Upon obtaining the generated images, we first perform object segmentation to extract precise masks before conducting quality evaluation. Object segmentation is performed using a combination of Segment Anything~\cite{kirillov2023segment} and Grounding DINO~\cite{liu2024grounding}. Segment Anything provides fine-grained segmentation masks for objects and regions of interest, generating pixel-level masks $M_t^{(n)}\in[0,1]^{H\times W}$ for each image $I_t^{(n)}$. Grounding DINO enables semantic grounding by associating text descriptions with corresponding image regions, facilitating the identification of semantically meaningful objects and their spatial locations. After object grounding and segmentation, 17618 instruction–image groups remain with valid object–mask pairs.

For each image with its corresponding segmentation mask, we perform quality evaluation using a multi-stage pipeline. First, we generate a detailed image caption $s_t^{(n)}$ for each image $I_t^{(n)}$ using a vision-language model, providing rich semantic descriptions of the image content. The image captions are then fed into a consistency evaluation module based on Qwen-VL-Max~\cite{wang2024qwen2}, which assesses the semantic consistency and coherence of the generated image set $\{I_t^{(n)}\}_{t=1}^{T^{(n)}}$ with respect to the instruction $c^{(n)}$. Images that fail to meet the consistency threshold are filtered out from the training dataset. During this process, 7982 groups passed the semantic and aesthetic evaluation stages.

Subsequently, an aesthetic evaluation module~\cite{schuhmann2022laion} assesses the visual quality of each image, identifying and removing images with poor aesthetic quality or visual artifacts. This ensures that the training dataset contains only high-quality images that contribute positively to model learning. Finally, a total of 7517 high-quality data groups are filtered to form our final training set.

\subsubsection{Annotation generation}
For each image that passes both consistency and aesthetic evaluation, we generate comprehensive text annotations. Using the image captions $s_t^{(n)}$ already generated during the quality evaluation stage, we retain the full image captions for the passed images. Additionally, for each segmented region identified by the masks $M_t^{(n)}$, we generate masked region descriptions $d_t^{(n)}$ that describe the content within the masked area, enabling fine-grained semantic understanding at both the image level and the region level.

The final training dataset consists of image groups $\{I_t^{(n)}\}_{t=1}^{T^{(n)}}$ paired with their corresponding masks $\{M_t^{(n)}\}_{t=1}^{T^{(n)}}$, image captions $\{s_t^{(n)}\}_{t=1}^{T^{(n)}}$, and masked region descriptions $\{d_t^{(n)}\}_{t=1}^{T^{(n)}}$. This comprehensive annotation enables the model to learn both semantic alignment and geometric consistency across image groups, supporting the training of GroupEditing's implicit-explicit correspondence fusion mechanism.

\begin{figure}[t]
    \centering
    \includegraphics[width=0.8\linewidth]{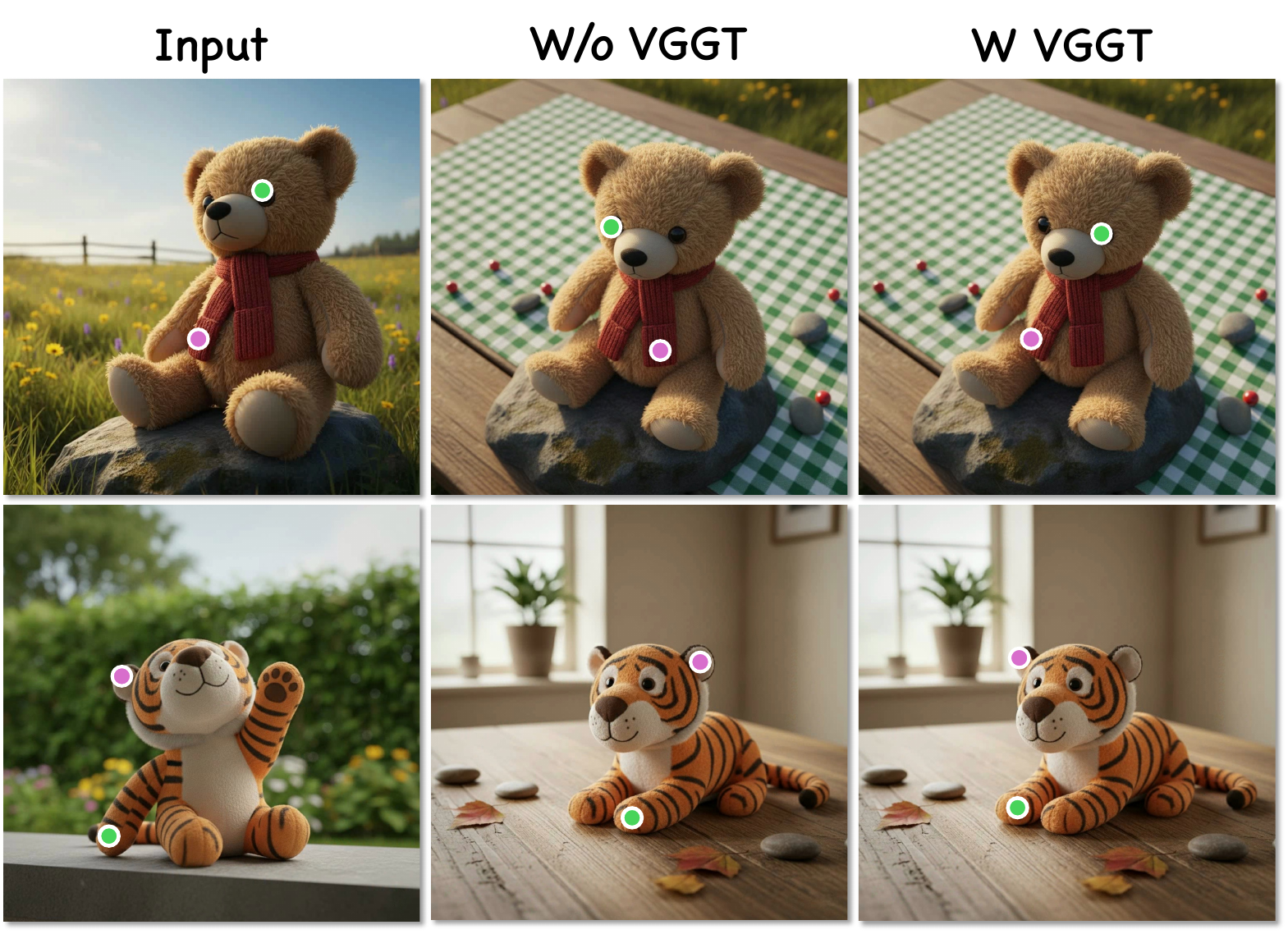}
    \caption{\textbf{Visual illustration about correspondence using VGGT.} Comparisons of the input and our correspondence prediction using VGGT.  The results without VGGT show less accurate alignment, while the inclusion of VGGT improves consistency and precision in the correspondence.}
    \label{fig:vggt_visual}
\end{figure}

\subsection{Group Editing}
\label{subsec:groupedit}

\subsubsection{Pipeline overview}
\label{subsubsec:pipeline}

As illustrated in the left part of Fig.~\ref{fig:qualitative-intro}, the pipeline processes an input image sequence $\{I_t\}_{t=1}^T$ through several stages. A fixed latent map $\mathcal{Z}:\mathbb{R}^{H\times W\times 3}\to\mathbb{R}^{C\times H'\times W'}$ encodes each frame: $z_t=\mathcal{Z}(I_t)$. At the scheduler parameter $\tau\in[0,1]$, noisy latents are constructed as:
\begin{equation}
  x_t(\tau) = \alpha(\tau)z_t + \sigma(\tau)\epsilon,\quad \epsilon\sim\mathcal{N}(0,\mathbf{I}).
\end{equation}
The latents are patchified with patch size $(p_t,p_h,p_w)$ into token sequences of length $S=\frac{T}{p_t}\frac{H'}{p_h}\frac{W'}{p_w}$. Text condition $c$ is encoded by a T5 encoder into contextual embeddings $u=\mathcal{E}_c(c)\in\mathbb{R}^{L\times D_c}$. 

For positional encoding, we apply Identity-RoPE to obtain the baseline 2D spatial positional map $\Pi_{\mathrm{Id}}(t,h,w)$ for the latent tokens, which provides precise pixel-level alignment within each image. When a displacement field $\Delta$ is available, we apply Ge-RoPE to obtain the geometry-aware positional map $\Pi_{\mathrm{Ge}}(t,h,w)$ that adjusts spatial indices according to $\Delta$.

Explicit dense tokens $G\in\mathbb{R}^{F\times H_g\times W_g\times D}$ are extracted from VGGT over the input set, with a positional map $\Pi_G$ defined on the $(F,H_g,W_g)$ grid. These tokens are augmented internally to the latent token sequence within each transformer block before self-interaction:
\begin{equation}
  \tilde{X} = [X_{\mathrm{latent}}, G],\quad \tilde{\Pi} = [\Pi, \Pi_G],
\end{equation}
where $X_{\mathrm{latent}}$ denotes the patchified latent tokens, $\Pi$ is either $\Pi_{\mathrm{Id}}$ or $\Pi_{\mathrm{Ge}}$ depending on the availability of $\Delta$, and $\Pi_G$ is the positional map for VGGT tokens obtained by mapping the backbone's per-axis frequency banks to the $(F,H_g,W_g)$ grid. The augmented sequence is processed through self-attention with the fused positional encoding $\tilde{\Pi}$, followed by cross-attention to $u$, and feed-forward layers. After self-attention, the output is truncated back to the original latent token dimensionality (only $X_{\mathrm{latent}}$ tokens are retained), ensuring that subsequent operations and integration remain in the latent space while benefiting from the cross-view constraints provided by $G$. The backbone predicts a velocity field:
\begin{equation}
  \hat{v}_\theta(x_t(\tau), c, u, \Pi, G, \Pi_G; \tau)\in\mathbb{R}^{C\times H'\times W'},
\end{equation}
which is integrated along the scheduler trajectory to obtain edited latents $\{\tilde{z}_t\}$, decoded $\mathcal{Z}^{-1}$ to produce images $\{\tilde{I}_t\}$.

\definecolor{mygreen}{RGB}{179, 200, 156}
\definecolor{myred}{RGB}{255, 179, 156}
\definecolor{myyellow}{RGB}{255, 228, 161}
\definecolor{myblue}{RGB}{172, 245, 255}

\begin{algorithm}[H]
\caption{Identity-RoPE for pixel alignment}
\label{alg:identity-rope}
\begin{algorithmic}[1]
\Require Base frequency $\theta \in \mathbb{R}_{>0}$, embedding dimensions $D_t, D_h, D_w \in \mathbb{N}$, number of images $T \in \mathbb{N}$, spatial dimensions $H', W' \in \mathbb{N}$, object masks $\{M_t\}_{t=1}^T$ where $M_t \in [0,1]^{H' \times W'}$
\Ensure Positional encoding $\Pi_{\mathrm{Id}} \in \mathbb{C}^{(T \times H' \times W') \times D/2}$ where $D = D_t + D_h + D_w$
\begin{tcolorbox}[colback=myred!60!white,
colframe=white, arc=3pt, outer arc=3pt,
boxrule=0pt, left=0pt, right=0pt,
top=0pt, bottom=0pt,
enlarge left by=-3pt] %
\For{$t = 1$ to $T$}
    \State $x_1^{(t)} \gets \min_{(h,w) \in \{M_t > 0.5\}} w$
    \State $y_1^{(t)} \gets \min_{(h,w) \in \{M_t > 0.5\}} h$
    \State $x_2^{(t)} \gets \max_{(h,w) \in \{M_t > 0.5\}} w$
    \State $y_2^{(t)} \gets \max_{(h,w) \in \{M_t > 0.5\}} h$
\EndFor
\end{tcolorbox}
\begin{tcolorbox}[colback=myyellow!60!white,
colframe=white, arc=3pt, outer arc=3pt,
boxrule=0pt, left=0pt, right=0pt,
top=0pt, bottom=0pt,
enlarge left by = -3pt] %
\For{$k \in \{h, w, t\}$}
    \State $\mathbf{d}_k \gets [0, 1, \ldots, D_k/2-1]^T \in \mathbb{N}^{D_k/2}$
    \State $\mathbf{f}_k \gets \theta^{-2\mathbf{d}_k/D_k} \in \mathbb{R}^{D_k/2}$
    \State $\mathbf{p}_k \gets [0, 1, \ldots, S_k-1]^T$ where $S_k \in \{H', W', T\}$
    \State $\Phi_k \gets \exp(\mathrm{i} \mathbf{p}_k \mathbf{f}_k^T) \in \mathbb{C}^{S_k \times D_k/2}$
\EndFor
\end{tcolorbox}
\begin{tcolorbox}[colback=mygreen!60!white,
colframe=white, arc=3pt, outer arc=3pt,
boxrule=0pt, left=0pt, right=0pt,
top=0pt, bottom=0pt,
enlarge left by = -3pt] %
\For{$t = 1$ to $T$}
    \For{$(h, w) \in [0, H'-1] \times [0, W'-1]$}
        \If{$x_1^{(t)} \leq w \leq x_2^{(t)} \wedge y_1^{(t)} \leq h \leq y_2^{(t)}$}
            \State $\tilde{h} \gets h - y_1^{(t)}$, $\tilde{w} \gets w - x_1^{(t)}$
        \Else
            \State $\tilde{h} \gets h$, $\tilde{w} \gets w$
        \EndIf
        \State $\Phi_h^{(t)}(h, w) \gets \Phi_h(\tilde{h})$, $\Phi_w^{(t)}(h, w) \gets \Phi_w(\tilde{w})$
    \EndFor
\EndFor
\end{tcolorbox}
% \begin{tcolorbox}[colback=myblue!60!white,
% colframe=white, arc=3pt, outer arc=3pt,
% boxrule=0pt, left=0pt, right=0pt,
% top=0pt, bottom=0pt,
% enlarge left by = -3pt] %
\State $\Phi_t \gets \Phi_t \otimes \mathbf{1}_{H'} \otimes \mathbf{1}_{W'} \in \mathbb{C}^{T \times H' \times W' \times D_t/2}$
\State $\Phi_h \gets \mathbf{1}_T \otimes \Phi_h \otimes \mathbf{1}_{W'} \in \mathbb{C}^{T \times H' \times W' \times D_h/2}$
\State $\Phi_w \gets \mathbf{1}_T \otimes \mathbf{1}_{H'} \otimes \Phi_w \in \mathbb{C}^{T \times H' \times W' \times D_w/2}$
\State $\Phi \gets \mathrm{concat}(\Phi_t, \Phi_h, \Phi_w) \in \mathbb{C}^{T \times H' \times W' \times D/2}$
\State $\Pi_{\mathrm{Id}} \gets \mathrm{reshape}(\Phi, (T \times H' \times W', D/2))$
% \end{tcolorbox}
\State \Return $\Pi_{\mathrm{Id}}$
\end{algorithmic}
\end{algorithm}

\subsubsection{Geometry-enhanced RoPE (Ge-RoPE)}
\label{subsubsec:ge-rope}

The top-right part of Fig.~\ref{fig:qualitative-intro}  and Fig.~\ref{fig:vggt_visual} illustrate Ge-RoPE, which improves alignment between VGGT features and latent features by incorporating spatial disparity information with confidence levels from displacement fields. Given a displacement field $\Delta(h,w)=(\Delta_h, \Delta_w) \in \mathbb{R}^{2 \times H' \times W'}$ that encodes pixel-level correspondences, confidence is derived from disparity magnitude (larger values indicate higher confidence), following a depth-disparity relationship similar to depth-based forward splatting~\cite{zhao2024stereocrafter}. The displacement field is resized to match latent resolution $(H', W')$ using bilinear interpolation, divided by the patch size (typically 16), and smoothed independently for $h$ and $w$ components using a 2D Gaussian kernel ($\mu=21$, $\sigma=11$) to prioritize high-confidence correspondences.

We construct warped spatial grids by adding the smoothed displacement to original grid indices: $\tilde{h}(h,w) = h + \Delta_h^{\mathrm{smooth}}(h,w)$ and $\tilde{w}(h,w) = w + \Delta_w^{\mathrm{smooth}}(h,w)$, clamped to $[0, H'-1]$ and $[0, W'-1]$ respectively. These warped grids index the precomputed frequency banks using nearest neighbor indexing, resulting in geometry-aware frequency tensors $\Phi_h^{\mathrm{Ge}}$ and $\Phi_w^{\mathrm{Ge}}$. The temporal component remains unchanged: $\Phi_t^{\mathrm{Ge}} = \Phi_t$. The final Ge-RoPE encoding is:
\begin{equation}
  \Pi_{\mathrm{Ge}} = \mathrm{concat}(\Phi_t, \Phi_h^{\mathrm{Ge}}, \Phi_w^{\mathrm{Ge}}) \in \mathbb{C}^{S_t \times S_h \times S_w \times D/2}.
\end{equation}
Applied to queries and keys through  multiplication:
\begin{equation}
  q' = \mathrm{Re}(q^{\mathbb{C}} \odot \Pi_{\mathrm{Ge}}),\quad k' = \mathrm{Re}(k^{\mathbb{C}} \odot \Pi_{\mathrm{Ge}}),
\end{equation}
where $\odot$ denotes element-wise complex multiplication, $q^{\mathbb{C}}, k^{\mathbb{C}}$ are complex representations of queries and keys (reshaped from real vectors), and $\mathrm{Re}(\cdot)$ extracts the real part. This warping aligns latent token positional encodings with the geometric structure in VGGT features.

\subsubsection{Identity-RoPE}
\label{subsubsec:identity-rope}

The bottom-right part of Fig.~\ref{fig:qualitative-intro} illustrates Identity-RoPE, which maintains identity consistency across the group through precise pixel-level alignment within each image via a 2D spatial positional encoding. Identity-RoPE applies separate 1D RoPE components for height and width using precomputed frequency banks. The frequencies are $\omega^{(i)}_h = \theta^{-2i/D_h}$ and $\omega^{(i)}_w = \theta^{-2i/D_w}$ for $i \in \{0, 1, \ldots, D_h/2-1\}$ and, $i \in \{0, 1, \ldots, D_w/2-1\}$ respectively, where $\theta$ is the base frequency. For spatial position $(h, w)$ where $h \in \{0, 1, \ldots, H'-1\}$ and $w \in \{0, 1, \ldots, W'-1\}$, the 1D RoPE components are:
\begin{equation}
\begin{split}
  \Phi_h(h) &= \exp(\mathrm{i} \boldsymbol{\omega}_h \cdot h) \in \mathbb{C}^{D_h/2}, \\
  \Phi_w(w) &= \exp(\mathrm{i} \boldsymbol{\omega}_w \cdot w) \in \mathbb{C}^{D_w/2},
\end{split}
\end{equation}
where $\boldsymbol{\omega}_h = [\omega^{(0)}_h, \omega^{(1)}_h, \ldots, \omega^{(D_h/2-1)}_h]^T$ and $\boldsymbol{\omega}_w = [\omega^{(0)}_w, \omega^{(1)}_w, \ldots, \omega^{(D_w/2-1)}_w]^T$ are frequency vectors, and $\mathrm{i}$ is the imaginary unit.

For robust identity alignment, we use a bounding rectangle-based matching strategy. For each image $t$, we compute the smallest bounding rectangle $\mathcal{R}_t = \{(x_1^{(t)}, y_1^{(t)}), (x_2^{(t)}, y_2^{(t)})\}$ from the segmentation mask $M_t$, where $(x_1^{(t)}, y_1^{(t)})$ and $(x_2^{(t)}, y_2^{(t)})$ are the bottom-left and top-right corners. Spatial coordinates are determined conditionally:
\begin{equation}
  (\tilde{h}, \tilde{w}) = \begin{cases}
    (h - y_1^{(t)}, w - x_1^{(t)}) & \text{if } (h, w) \in \mathcal{R}_t \\
    (h, w) & \text{otherwise}
  \end{cases}
\end{equation}
where $\mathcal{R}_t = \{(h, w) : x_1^{(t)} \leq w \leq x_2^{(t)} \wedge y_1^{(t)} \leq h \leq y_2^{(t)}\}$. Pixels within $\mathcal{R}_t$ use normalized coordinates relative to the rectangle's origin, ensuring corresponding object regions share identical positional encodings regardless of absolute positions. The complete positional encoding is:
\begin{equation}
  \Pi_{\mathrm{Id}}(t, h, w) = \mathrm{concat}(\Phi_t(t), \Phi_h(\tilde{h}), \Phi_w(\tilde{w})) \in \mathbb{C}^{D/2},
\end{equation}
where $\Phi_t(t)$ is the temporal encoding component, and $D = D_t + D_h + D_w$. Applied to queries and keys:
\begin{equation}
  q' = \mathrm{Re}(q^{\mathbb{C}} \odot \Pi_{\mathrm{Id}}),\quad k' = \mathrm{Re}(k^{\mathbb{C}} \odot \Pi_{\mathrm{Id}}),
\end{equation}
where $\odot$ denotes element-wise complex multiplication. This ensures tokens within corresponding object regions share consistent positional signatures, as visualized in the bottom-right part of Fig.~\ref{fig:qualitative-intro}.

\begin{figure}[t]
    \centering
    \includegraphics[width=0.8\linewidth]{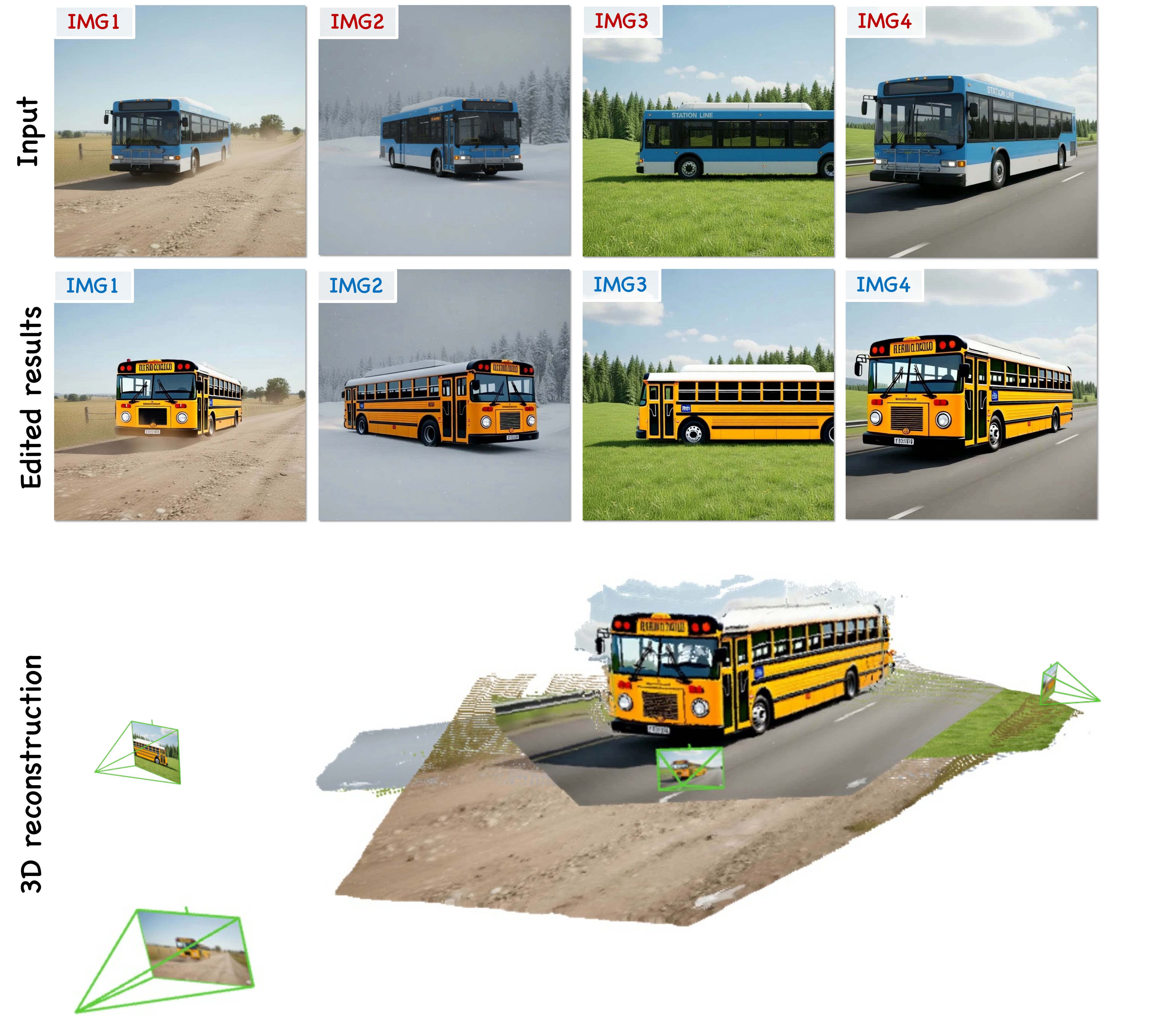}
    \caption{\textbf{3D reconstruction using editing results.} The top row shows input images, the middle row displays the edited results, and the bottom row demonstrates 3D reconstruction using Must3R, aligning 2D points to create a 3D model of the modified bus.}
    \label{fig:app_3d}
\end{figure}

\begin{figure}[t]
    \centering
    \includegraphics[width=0.8\linewidth]{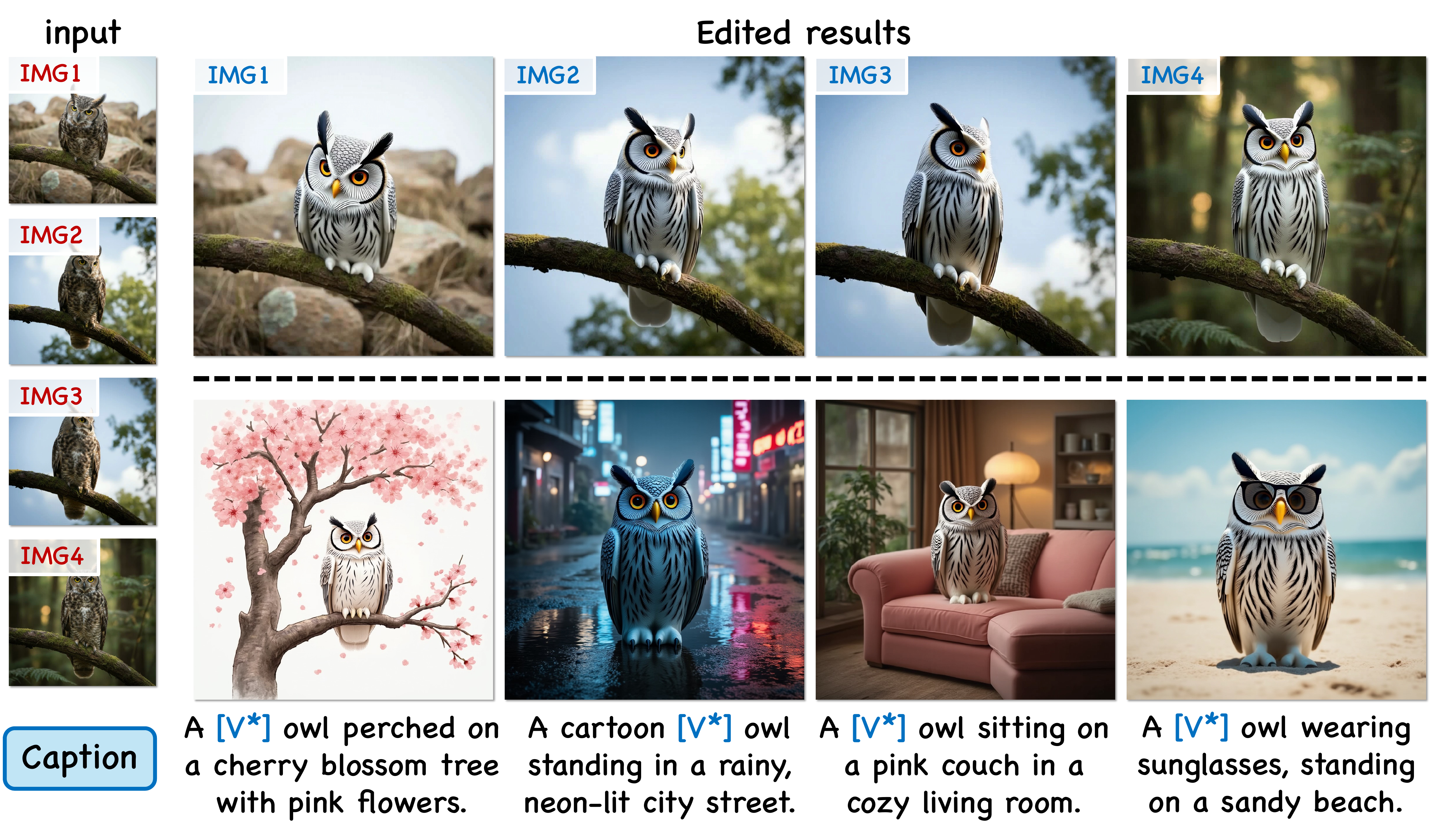}
    \caption{\textbf{Image customization using editing results.} Customized generation is achieved by embedding the edited concepts into the generative model, leveraging the outputs from our consistent editing method.}
    \label{fig:app_customization}
\end{figure}

%% file: sec/4_experiment.tex
\section{Experiment}
\label{sec:experiment}

\subsection{Implementation Details}

We train our model using on WAN-2.1~\cite{wan2025}, a transformer-based video diffusion model~\cite{seawead2025seaweed,kong2024hunyuanvideo,HaCohen2024LTXVideo}.
The optimization is performed using AdamW~\cite{loshchilov2017decoupled} with a weight decay of 0.01 and an initial learning rate of $1\times10^{-4}$. Training is conducted at a spatial resolution of $528 \times 528$ and a batch size of 8 on 8 NVIDIA A800 GPUs under PyTorch. 
More implementation details and quantitative metrics are provided in the supplementary materials.

\begin{figure*}[ht]
    \centering
    \includegraphics[width=0.88\textwidth]{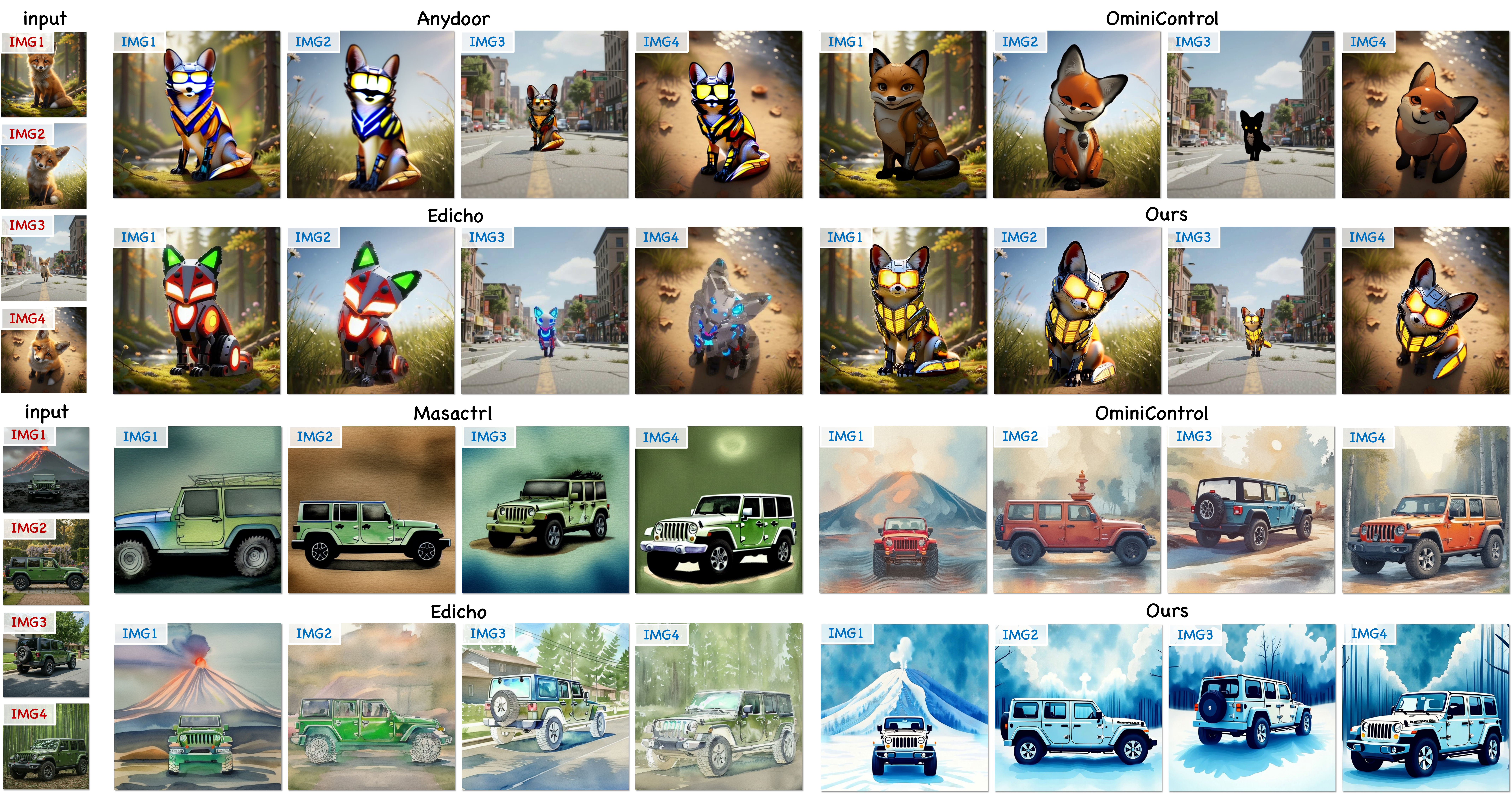}
    \caption{\textbf{Visual comparison with SOTA image editing on local and global editing }.The top row shows results for different methods on local edits. The bottom row presents our method alongside other techniques for global editing. Our approach demonstrates superior consistency and high-quality edits across both local and global changes. The text prompts are \enquote{\textit{A cartoon fox with futuristic robotic armor and orange details.}} and \enquote{\textit{A jeep vehicle in outdoor settings, including snow-covered style, rendered in soft watercolor painting style.}}.
    }
    \label{fig:visual_compare_local}
\end{figure*}

\begin{figure}[t]
    \centering
    \includegraphics[width=0.95\linewidth]{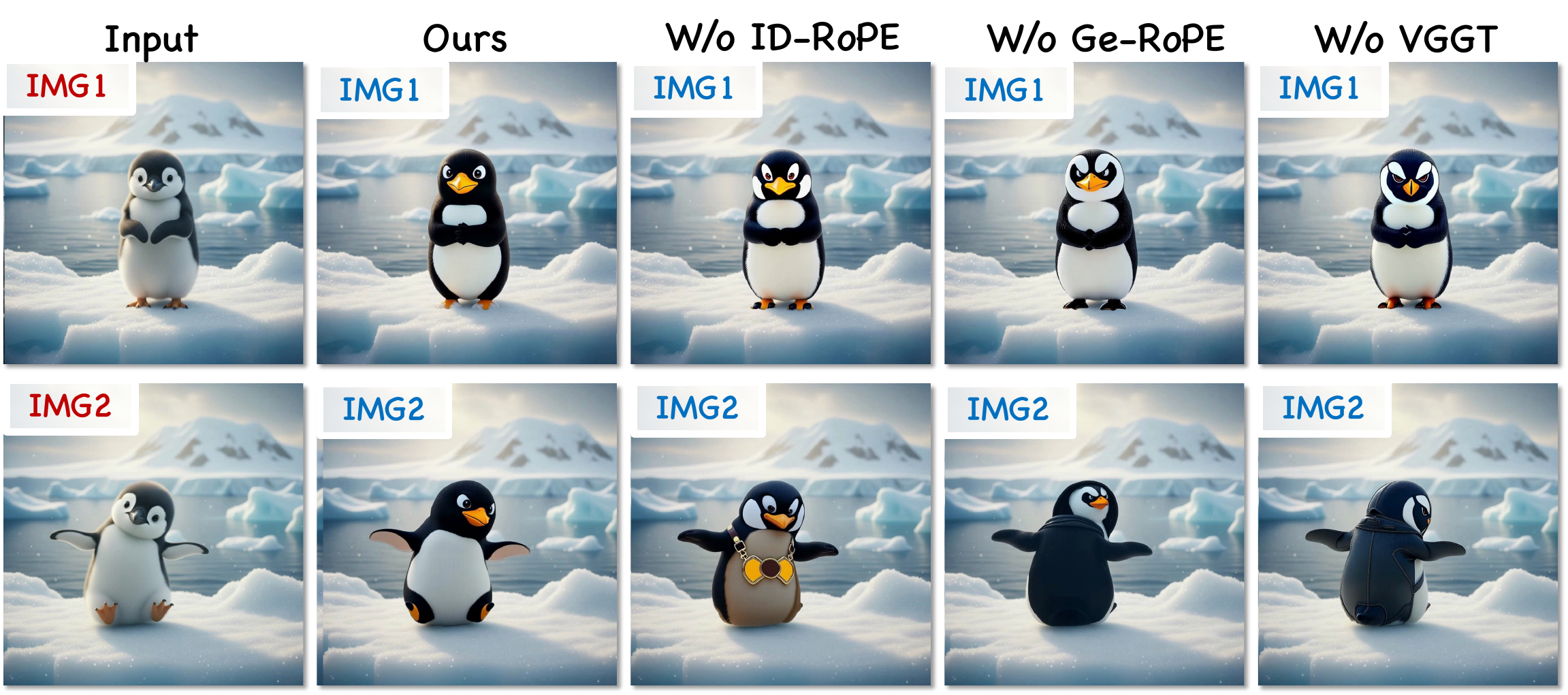}
    \caption{\textbf{Visual ablation about proposed modules.}  We provide a visual ablation study on the effectiveness of various mechanisms. The prompt is \enquote{\textit{The penguin has a black and white body with orange feet and beak.}}}
    \label{fig:ablation}
\end{figure}

\begin{table*}[t]
  \centering
  \caption{\textbf{Comparison with state-of-the-art image editing on local editing}.   \textcolor{Red}{\textbf{Red}} and \textcolor{Blue}{\textbf{Blue}} denote the best and second best results.} 
  \resizebox{0.95\textwidth}{!}{%
    \begin{tabular}{l|ccccc|cccc}
      \toprule
      \multirow{2}{*}{Method}
        & \multicolumn{5}{c|}{Quantitative metrics}
        & \multicolumn{4}{c}{User study} \\
      \cmidrule(lr){2-6}\cmidrule(lr){7-10}
        & CLIP-Score↑ & Aesthetic-Score↑ & DINO-Score↑ & Editing Consis. ↑   & PSNR↑ & Iden. Consis.↓ & Aesthetic.↓ & App. Fidelity↓ & Overall↓ \\
      \midrule
      \midrule
      % \midrule \\
      % \midrule
        % Adobe Firefly~\cite{xiao2024video}
        % &  0.2297 & 0.6511 & 0.9797 
        % &  0.9593 & 0.9413 & 0.4581 & 0.9716 \\
        Anydoor~\cite{chen2024anydoor}
        &  0.2728 & 4.72 & 0.7208  &  0.8697
        & 0.6182 & 3.56 & 3.23 & 3.60 & 3.32\\
        OminiControl~\cite{tan2025ominicontrol}
        &  0.2902 & \textbf{\textcolor{Blue}{5.10}} & 0.7326 &  0.8676
        & 0.6457 & 2.84 & 2.84 & \textbf{\textcolor{Blue}{1.96}} & 2.66 \\
      Edicho~\cite{bai2025edicho}
        & \textbf{\textcolor{Blue}{0.3059}} &  4.89& \textbf{\textcolor{Blue}{0.8080}} & \textbf{\textcolor{Blue}{0.8988}} 
        & \textbf{\textcolor{Blue}{0.6935}} & \textbf{\textcolor{Blue}{1.93}} & \textbf{\textcolor{Blue}{2.47}} & 2.93 & \textbf{\textcolor{Blue}{1.95}}\\
      \textbf{Ours}
        & \textbf{\textcolor{Red}{0.3122}} & \textbf{\textcolor{Red}{5.39}} & \textbf{\textcolor{Red}{0.8168}} & \textbf{\textcolor{Red}{0.9239}}
        & \textbf{\textcolor{Red}{0.7624}} & \textbf{\textcolor{Red}{1.67}} & \textbf{\textcolor{Red}{1.46}} & \textbf{\textcolor{Red}{1.50}} & \textbf{\textcolor{Red}{1.47}}  \\
      \bottomrule
    \end{tabular}%
}
\label{tab:comparison_localediting}
\end{table*}

\begin{table}[t]
  \centering
  \caption{\textbf{Comparison with state-of-the-art image editing on global editing}. \textcolor{Red}{\textbf{Red}} and \textcolor{Blue}{\textbf{Blue}} denote the best and second best results, respectively.}
  \vspace{-5pt}
  \label{tab:comparison_global}
  \resizebox{\linewidth}{!}{%
    \begin{tabular}{l|cccc}
      \toprule
      \multirow{2}{*}{Method}
        & \multicolumn{4}{c}{Quantitative metrics} \\
      \cmidrule(lr){2-5}
        & CLIP-Score↑ & Aesthetic-Score↑ & DINO-Score↑ & Editing Consis.↑  \\
      \midrule
      % Masactrl~\cite{cao2023masactrl}
      %   & 0.2848 & 4.6524 & - & - & 0.8383 \\
      StyleAligned~\cite{hertz2024style}
        & 0.2910 & 4.75  & 0.8645 & 0.8714 \\
      % Cross-Image-Attention~\cite{alaluf2024cross}
      %   & - & - & - & - & -\\
      Edicho~\cite{bai2025edicho}
        &  \textbf{\textcolor{Blue}{0.2920}} & 4.59  & \textbf{\textcolor{Blue}{0.8714}} & 0.8859 \\
        OminiControl~\cite{tan2025ominicontrol}
        &  0.2918 & \textbf{\textcolor{Blue}{4.94}}  & 0.8553 & \textbf{\textcolor{Blue}{0.8965}} \\
      \textbf{Ours}
        &  \textbf{\textcolor{Red}{0.2987}} & \textbf{\textcolor{Red}{5.48}} & \textbf{\textcolor{Red}{0.9287}} & \textbf{\textcolor{Red}{0.9147}} \\
      \bottomrule
    \end{tabular}%
  }
\end{table}

\subsection{Applications}

\noindent\textbf{Image customization based on the consistent results.} To demonstrate the practical application of our consistent editing method, we use DreamBooth~\cite{ruiz2023dreambooth} and Low-Rank Adaptation (LoRA)~\cite{hu2022lora} for customized image generation. By leveraging the edited outputs from our method, we finetune a generative model with DreamBooth for 600 steps.  As shown in Fig.~\ref{fig:app_customization}, the fine-tuned model successfully generates images based on the edited images, enabling both novel concept generation and editing.

\noindent\textbf{3D reconstruction based on the consistent results.}
Our method also benefits 3D reconstruction by leveraging the Must3R~\cite{cabon2025must3r} to predict accurate 3D scene representations from consistent image pairs. Using the edited images as inputs, Must3R generates 3D point-based models and 2D matchings without requiring additional inputs like camera parameters. As shown in Fig.~\ref{fig:app_3d}, the reconstruction results confirm the editing consistency for the four sets of edits.

\subsection{Comparison with baselines.}

\noindent\textbf{Qualitative comparison.}
We present a qualitative evaluation of our approach with recent open-sourced state-of-the-art methods in local image editing. We compare our method with open-sourced approaches such as Anydoor~\citep{chen2024anydoor}, Paint-by-Example~\citep{yang2023paint}, OminiControl~\citep{tan2025ominicontrol}, and Edicho~\citep{bai2025edicho}. The visual results are shown in Fig.~\ref{fig:visual_compare_local}. 
% While these open-sourced methods are capable of editing images locally, they fail to maintain the editing consistency between multiple images. 
Anydoor~\cite{chen2024anydoor}, OminiControl~\cite{tan2025ominicontrol}, and Edicho~\cite{bai2025edicho} struggle to edit multiple images coherently. Our framework enables editing the image group consistently.
Additionally, we show the visual comparison with Masactrl~\cite{cao2023masactrl}, StyleAligned~\cite{hertz2024style}, Edicho~\cite{bai2025edicho}, and OminiControl~\cite{tan2025ominicontrol} in global image editing. As shown in Fig.~\ref{fig:visual_compare_local}, our approach demonstrates superior consistency and alignment across the multiple editing results.

% Our approach, on the other hand, excels in this aspect by offering both movement and accurate orientation control. 

%%here

\begin{table}[t]
  \centering
  \caption{\textbf{Quantitative ablation}. \textcolor{Red}{\textbf{Red}} and \textcolor{Blue}{\textbf{Blue}} denote the best and second best results, respectively.} 
  \label{tab:ablation}
  \resizebox{\linewidth}{!}{%
    \begin{tabular}{l|cccc}
      \toprule
      \multirow{2}{*}{Method}
        & \multicolumn{4}{c}{Quantitative metrics} \\
      \cmidrule(lr){2-5}
        & CLIP-Score↑ & Aesthetic-Score↑ & DINO-Score↑ & Editing Consis.↑  \\
      \midrule
      W/o VGGT~\cite{hertz2024style}
        & 0.2728 & 4.72 & 0.7208 & 0.8616  \\
      W/o Ge-RoPE~\cite{bai2025edicho}
        & \textbf{\textcolor{Blue}{0.2902}} & 4.89 & \textbf{\textcolor{Blue}{0.7326}} & 0.8697 \\
      W/o Identity-RoPE~\cite{tan2025ominicontrol}
        & 0.2902 & \textbf{\textcolor{Blue}{4.89}} & 0.7326 & \textbf{\textcolor{Blue}{0.9108}} \\
      \textbf{Ours}
        & \textbf{\textcolor{Red}{0.3122}} & \textbf{\textcolor{Red}{5.39}} & \textbf{\textcolor{Red}{0.8168}} & \textbf{\textcolor{Red}{0.9239}} \\
      \bottomrule
    \end{tabular}%
  }
\end{table}
% The comparisons are shown in Fig.~\ref{}

\noindent\textbf{Quantitative comparison.}
We compare our method with state-of-the-art image editing on the collected benchmark, GroupEditBench. It includes 800 image sets generated by a powerful T2I diffusion model. 
The GroupEditBench includes diverse content, including objects, humans, animals, and landscapes, and various styles of images, such as sketch and cyberpunk style. We split them into local and global image editing. For local image editing, we compare our framework with Anydoor~\cite{chen2024anydoor}, OminiControl~\cite{tan2025ominicontrol}, Edicho~\cite{bai2025edicho}. for global image editing, StyleAligned~\cite{hertz2024style}, OminiControl~\cite{tan2025ominicontrol} and Edicho~\cite{bai2025edicho} are considered to fair comparsion.  We evaluate these methods using several standard metrics (shown in Tab.~\ref{tab:comparison_localediting} and Tab.~\ref{tab:comparison_global}). For \enquote{Editing Consis.}, following Edicho, we use the feature similarity of the edited results to evaluate the editing consistency. We leverage the Laion-Aesthetic Score Predictor~\cite{schuhmann2022laion} for Aesthetic-Score. Additionally, we invited 20 volunteers to rank methods across four aspects, including identity consistency, aesthetic, appearance fidelity, and overall quality, on a 1 (best) to 4 scale. The average rank (lower is better) is shown in Tab.~\ref{tab:comparison_localediting}. Our method achieves the top result in both automatic metrics and human preference.

\subsection{Ablation study}

\noindent\textbf{Effectiveness of  Geometry-enhanced RoPE.}
We ablate various geometry-enhanced RoPE (Relative Positional Encoding) configurations during training. The visual results are provided in Fig.~\ref{fig:ablation}. As the geometry-enhanced RoPE is applied, we observe the improvements in the model’s spatial awareness and the accuracy of geometric transformations. Additionally, as shown in Tab.~\ref{tab:ablation}, we report the quantitative ablation results to further validate the effectiveness of geometry-enhanced RoPE.

\noindent\textbf{Effectiveness of Identity-RoPE.}
Thanks to our Identity-RoPE, the model can maintain a more consistent and stable representation of the object’s identity. In Fig.~\ref{fig:ablation}, we observe that the object’s identity remains significantly more consistent when equipping the Identity-RoPE layers. As shown in Tab.~\ref{tab:ablation}, our experiments quantitatively validate the benefit of incorporating Identity-RoPE in preserving identity consistency across different images.

%% file: sec/5_conclusion.tex
\section{Conclusion}
\label{sec:conclusion}
In this paper, we present GroupEditing, a novel framework designed to address the challenges of consistent multiple-image editing. By leveraging explicit geometric correspondences from VGGT with implicit priors from pre-trained video models, our method not only ensures accurate modifications but also maintains identity preservation across diverse images through the alignment-enhanced RoPE module. 
To support large-scale training, the framework is backed by GroupEditData, a new dataset that provides high-quality image groups with precise segmentation masks and detailed image captions.
Additionally, we evaluate GroupEditing using GroupEditBench, a comprehensive benchmark specifically designed to assess group-level editing performance.
Extensive experimental results demonstrate that GroupEditing outperforms SOTA methods, including visual quality, editing consistency, and semantic alignment, establishing it as a promising solution for a wide range of multi-image editing applications. 

% The proposed framework is supported by GroupEditData, a new dataset tailored for large-scale training, and evaluated using GroupEditBench, a benchmark that measures group-level editing performance. Extensive experiments show that GroupEditing outperforms state-of-the-art techniques in visual quality, editing consistency, making it a promising solution for multi-image editing tasks.

%% file: sec/X_suppl.tex
\clearpage
\setcounter{page}{1}
\maketitlesupplementary

\tableofcontents
\setcounter{page}{1}
\renewcommand{\thepage}{S\arabic{page}}
\setcounter{section}{0}
\setcounter{figure}{0}
\setcounter{table}{0}

\appendix

\section{Rationale}
\label{sec:rationale}
Having the supplementary compiled together with the main paper means that:
\begin{itemize}
\item The supplementary can back-reference sections of the main paper, for example, we can refer to introduction;
\item The main paper can forward reference sub-sections within the supplementary explicitly (\textit{e.g.} referring to a particular experiment); 
\item When submitted to arXiv, the supplementary will already included at the end of the paper.
\end{itemize}
To split the supplementary pages from the main paper, you can use \href{https://support.apple.com/en-ca/guide/preview/prvw11793/mac#:~:text=Delete%20a%20page%20from%20a,or%20choose%20Edit%20%3E%20Delete).}{Preview (on macOS)}, \href{https://www.adobe.com/acrobat/how-to/delete-pages-from-pdf.html#:~:text=Choose%20%E2%80%9CTools%E2%80%9D%20%3E%20%E2%80%9COrganize,or%20pages%20from%20the%20file.}{Adobe Acrobat} (on all OSs), as well as \href{https://superuser.com/questions/517986/is-it-possible-to-delete-some-pages-of-a-pdf-document}{command line tools}.

\section{Realted Work}
\noindent\textbf{Generative models for image editing.}
Image editing has seen remarkable progress driven by diffusion-based generative models~\cite{ho2020denoising,rombach2022high,brooks2023instructpix2pix,mokady2023null,cao2023masactrl,wang2025multishotmaster,tumanyan2023plug,chen2024zero,chen2024anydoor,winter2024objectdrop, z1,z2,z3,z17,z18, chen2025s2guidancestochasticselfguidance}.
Existing techniques can be broadly divided into inference‐time zero‐shot methods that edit images by manipulating the diffusion process itself (such as PnP~\cite{tumanyan2023plug}, Prompt2Prompt~\cite{mokady2023null}, and MasaCtrl~\cite{cao2023masactrl}), and training‐based methods, which achieve editing by fine‐tuning latent diffusion models (represented by ControlNet~\cite{zhang2023adding} and T2I‐Adapter~\cite{mou2024t2i}).
However, these methods remain tailored to single‐image editing. When applied to a group of related images, they often fail to maintain coherence in appearance and structure, resulting in inconsistencies. Existing efforts to enforce consistency, whether by propagation~\cite{winter2024objectdrop} or attention‐based correspondences~\cite{mokady2023null,cao2023masactrl}, are limited to small inputs and break down under complex geometric variation, in part due to the scarcity of suitable paired training data.
In response, we formalize the problem of \textit{Group‐Image Editing} and introduce \textit{\modelname}: a trainable framework that views related images as pseudo video frames to inherit implicit consistency priors from video models, while additionally incorporating an explicit correspondence module to ensure reliable alignment.

\begin{figure*}[ht]
    \centering
    \includegraphics[width=1.\textwidth]{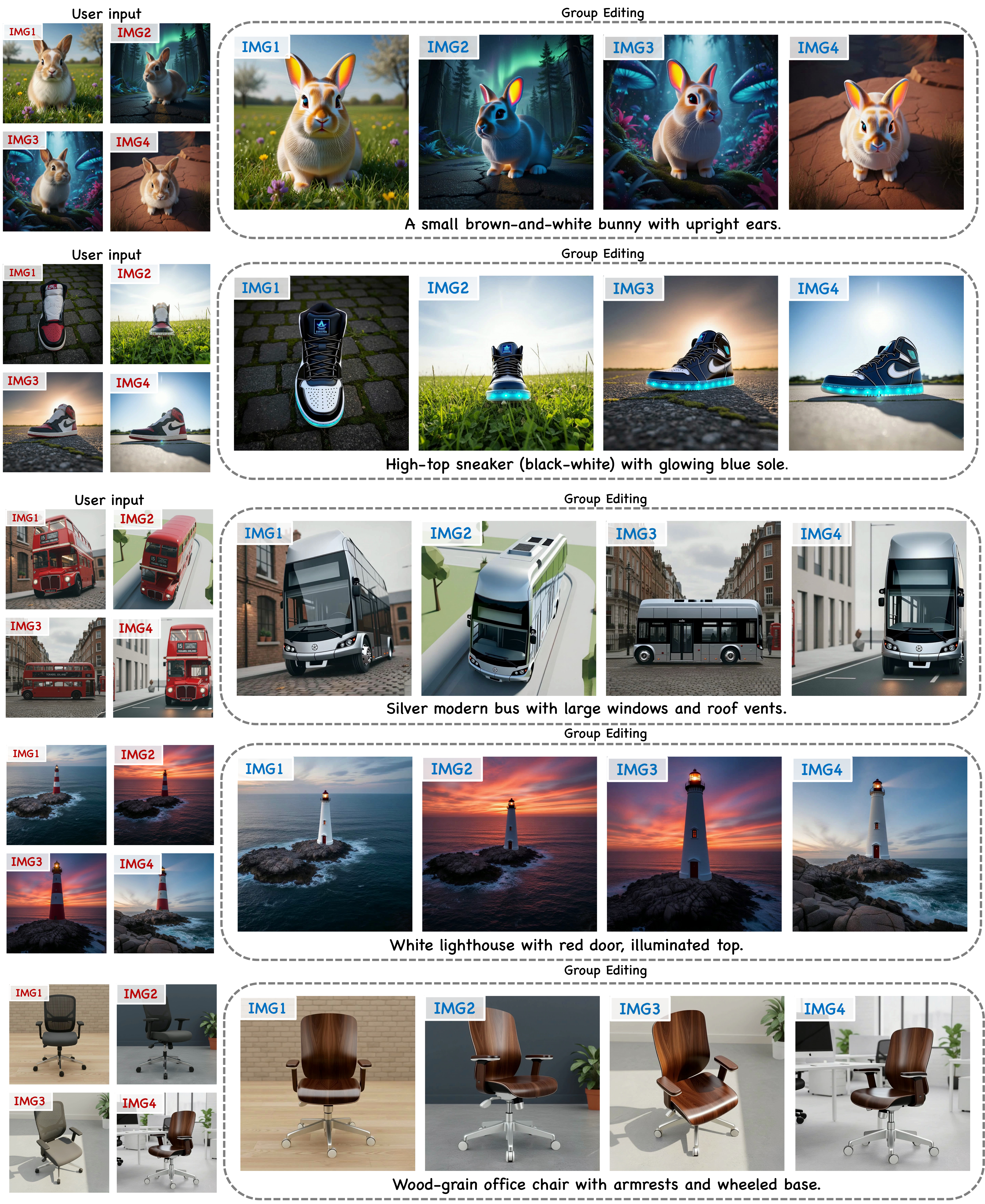}
    \caption{\textbf{More cases}. 
    }
    \label{fig:case1}
\end{figure*}

\begin{figure*}[ht]
    \centering
    \includegraphics[width=1.\textwidth]{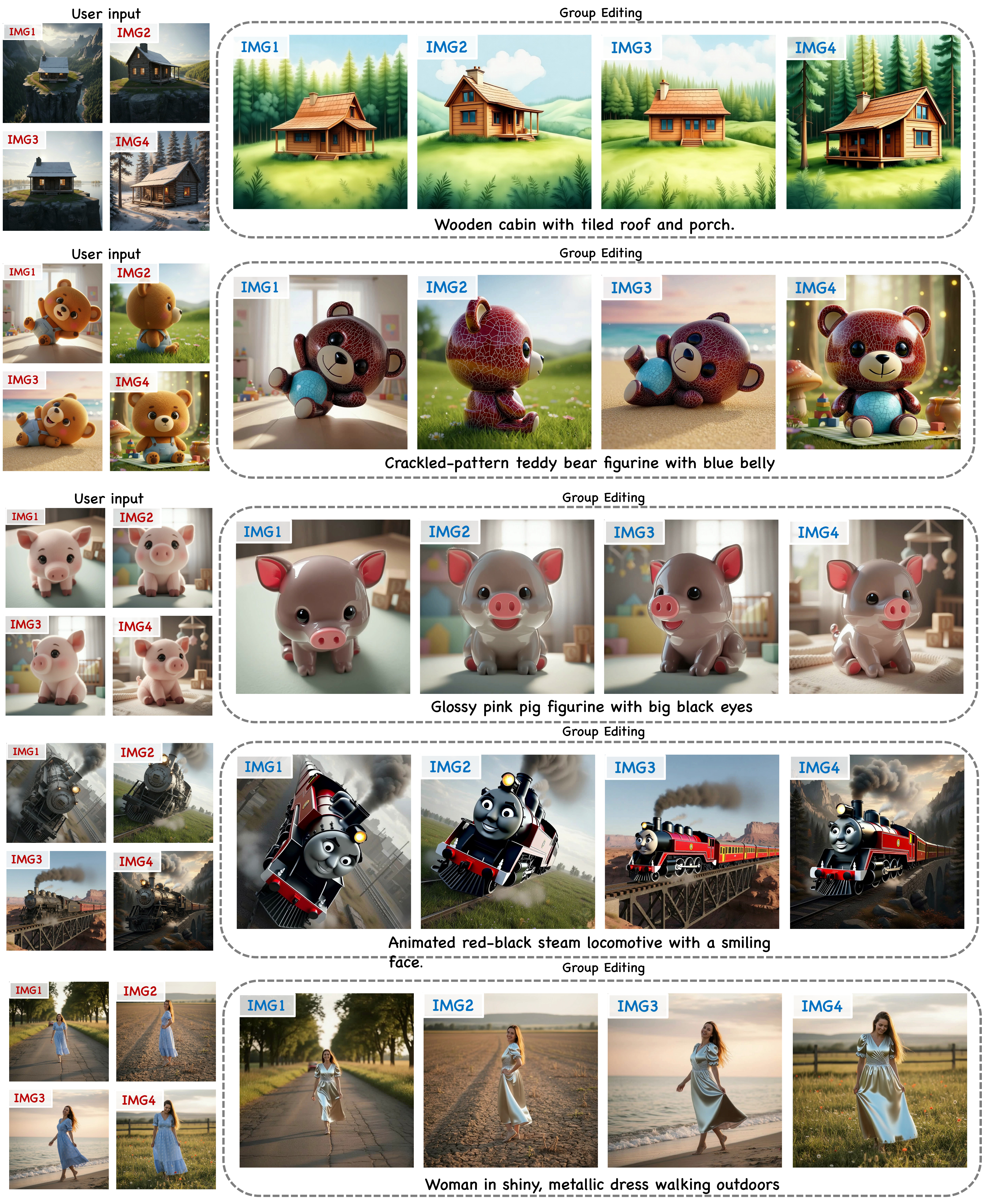}
    \caption{\textbf{More cases}. 
    }
    \label{fig:case2}
\end{figure*}

\begin{figure*}[ht]
    \centering
    \includegraphics[width=1.\textwidth]{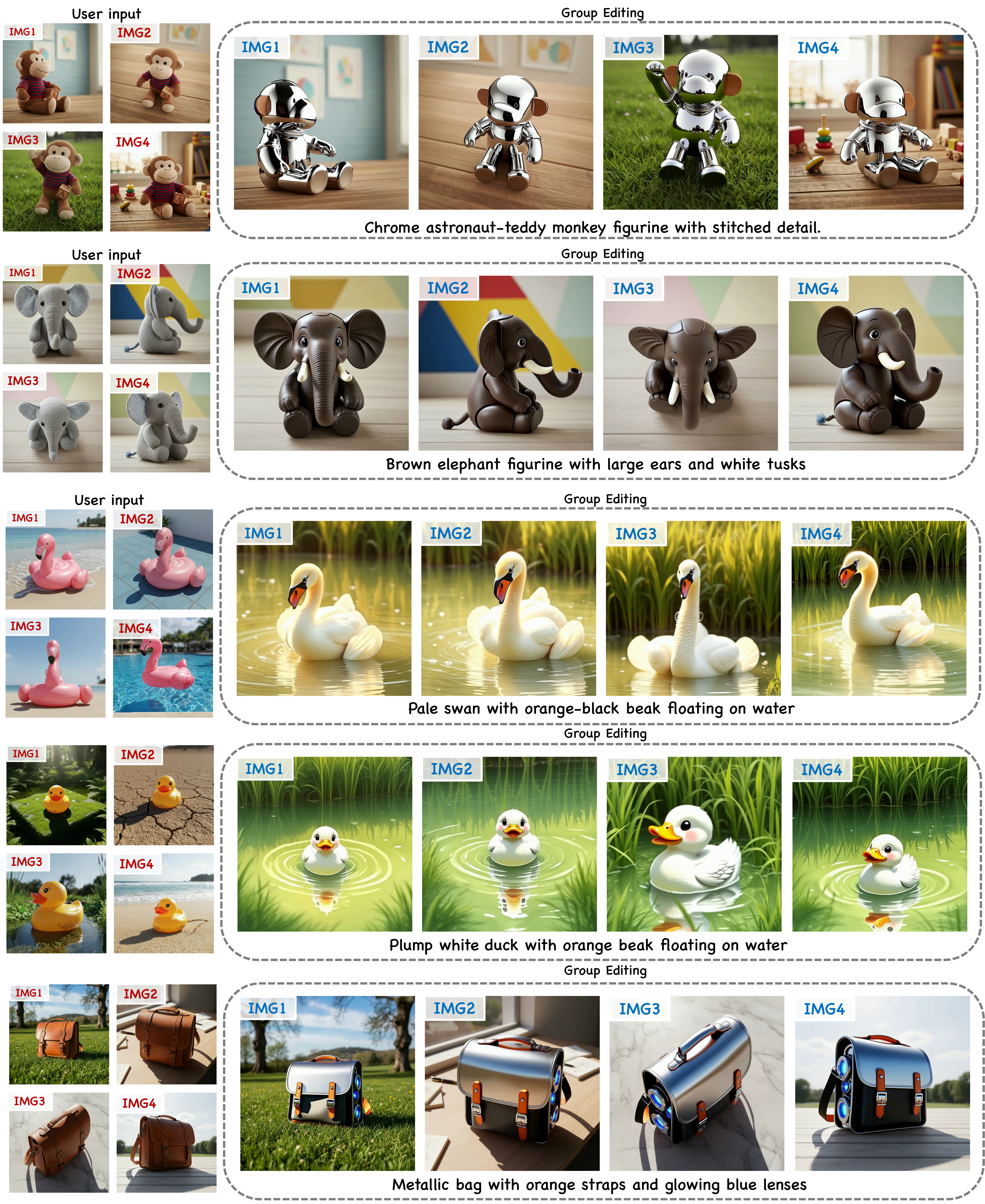}
    \caption{\textbf{More cases}. 
    }
    \label{fig:case3}
\end{figure*}

\begin{figure*}[ht]
    \centering
    \includegraphics[width=1.\textwidth]{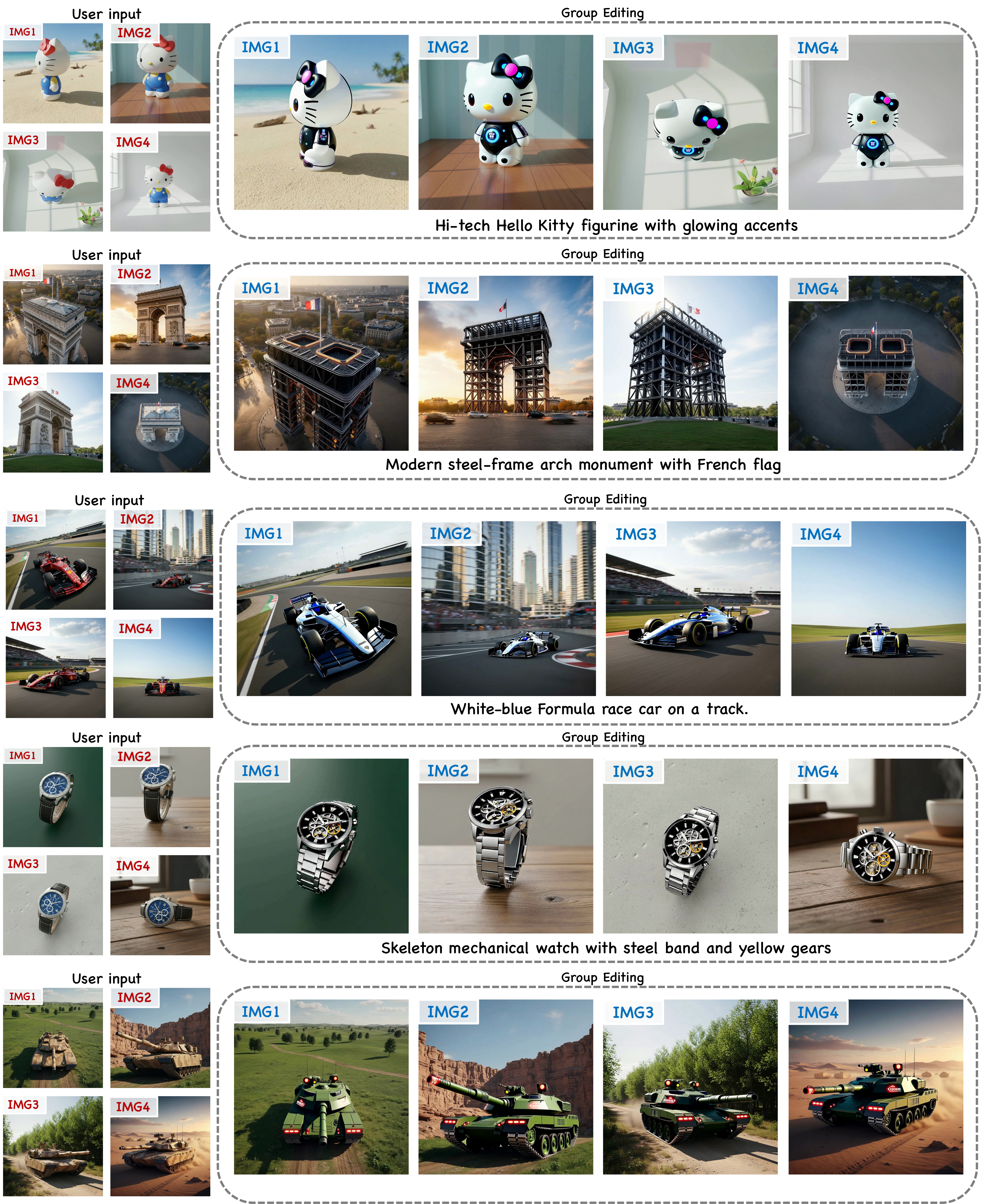}
    \caption{\textbf{More cases}. 
    }
    \label{fig:case4}
\end{figure*}

\begin{figure*}[ht]
    \centering
    \includegraphics[width=1.\textwidth]{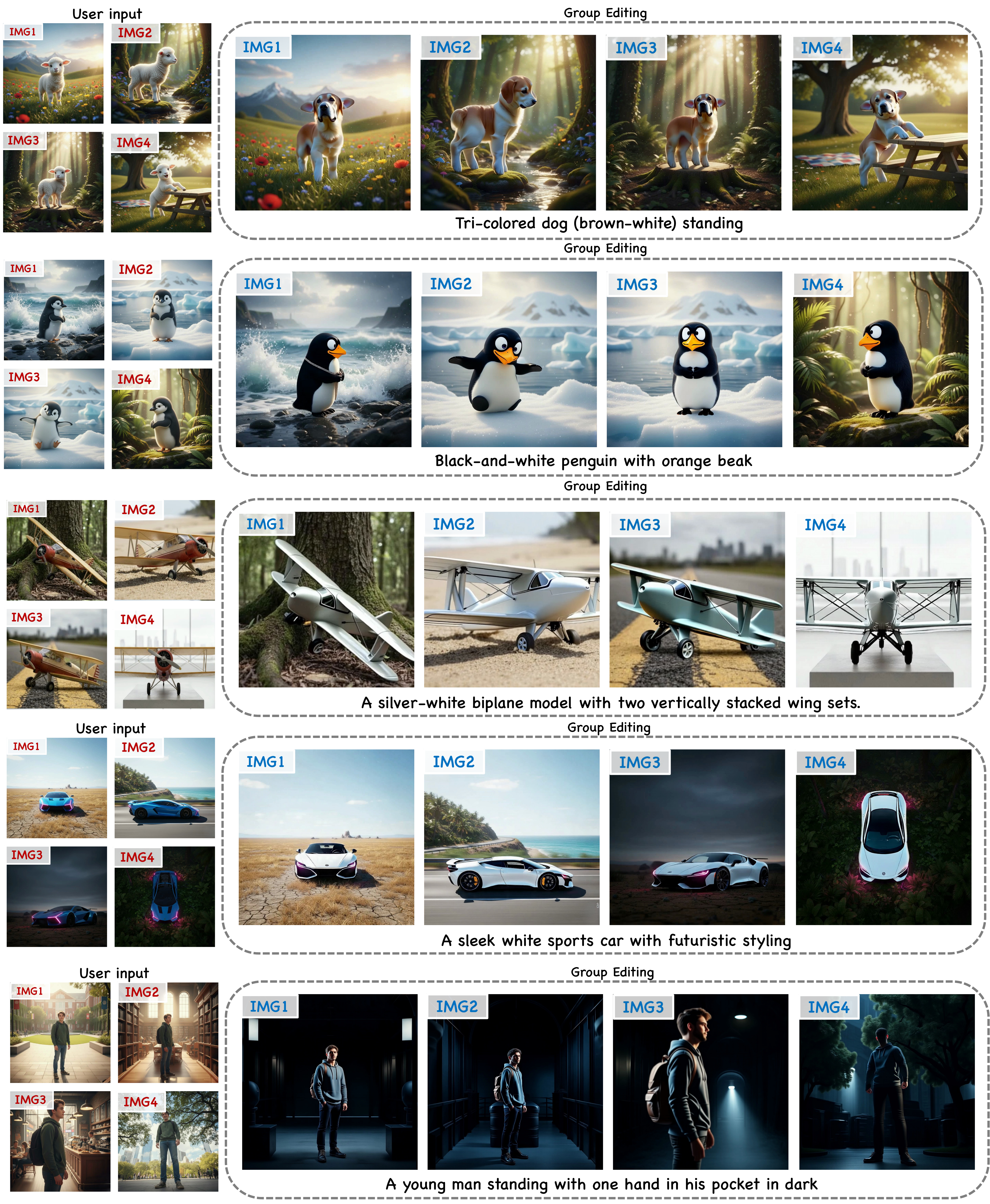}
    \caption{\textbf{More cases}. 
    }
    \label{fig:case5}
\end{figure*}

\noindent\textbf{Video prior for editing task.}
Video generative models~\cite{chen2025taming, zhang2025easycontrol, zhang2024ssr, song2025layertracer, song2025makeanything} provide powerful temporal consistency priors that can be effectively leveraged for image editing. Existing studies generally follow two directions: utilizing video data for training data curation and leveraging video models for inference-time guidance.
For the former, Bagel~\cite{bagel}, UniReal~\cite{chen2025unireal}, and OmniGen~\cite{xiao2025omnigen} sample temporally coherent frames from video data to create high-quality training sequences, a strategy that implicitly injects structural and appearance continuity into the resulting image models.
For the latter, Frame2Frame~\cite{rotstein2025pathways} utilizes a pre-trained video diffusion model to synthesize a sequence of frames and select an intermediate frame, thereby enforcing temporal smoothness and structural continuity directly during inference. 
ChronoEdit~\cite{wu2025chronoedit} leverages pretrained video generative models to reframe image editing as a video generation task, using the input and target images as video endpoints.
While effective for enhancing single-image quality or ensuring short-range consistency, these approaches do not solve the fundamental challenge of Group-Image Editing across diverse, static views. Our approach differs by re-framing the image group as a pseudo video sequence, allowing us to explicitly inherit the powerful spatio-temporal coherence and geometric priors of large-scale video models for robustly unified editing.

\section{More cases}

\section{Demo}
We provide the \textcolor{blue}{demo video} and \textcolor{blue}{project page}  in the file, please watch it for better illustration. 